\documentclass[fleqn,10pt]{wlscirep}
\usepackage[utf8]{inputenc}
\usepackage[T1]{fontenc}
\usepackage[square, numbers, sort&compress]{natbib}
\usepackage{tabularx} 
\usepackage{multirow} 
\usepackage{adjustbox}
\usepackage{algorithm}
\usepackage{algpseudocode}
\usepackage{amsmath}
\usepackage{booktabs}
\usepackage[colorlinks=true,allcolors=blue]{hyperref}
\usepackage{siunitx}
\usepackage{threeparttable}
\usepackage{caption}
\usepackage{graphicx}
\usepackage{calc} 
\usepackage{hyperref}
\usepackage{scrextend}
\newcommand{\semiSmall}{\fontsize{9.5pt}{10}\selectfont}

\renewenvironment{thebibliography}[1]{%
  \begin{oldthebibliography}{#1}%
  \semiSmall 
}{%
  \end{oldthebibliography}%
}
\title{Fairness and bias correction in machine learning for depression prediction: results from four study populations}

\author[1,*]{Vien Ngoc Dang}
\author[1]{Anna Cascarano}
\author[2]{Rosa H. Mulder}
\author[3]{Charlotte Cecil}
\author[4]{Maria A. Zuluaga}
\author[5]{Jerónimo Hernández-González}
\author[1]{Karim Lekadir}
\affil[1]{Departament de Matemàtiques i Informàtica, Facultat de Matemàtiques i Informàtica, Universitat de Barcelona, Spain}
\affil[2]{Department of Pediatrics, Erasmus University Rotterdam, the Netherlands
}
\affil[4]{Department of Child and Adolescent Psychiatry/Psychology, ErasmusMC-Sophia, Rotterdam, the Netherlands
}
\affil[4]{Data Science Department, EURECOM, Sophia Antipolis, France}
\affil[5]{Departament d'Informàtica, Matemàtica Aplicada i Estadística, Universitat de Girona, Spain
}
\affil[*]{dangn@ub.edu}

\keywords{Machine learning for depression prediction, Algorithmic fairness, Bias mitigation, Novel post-hoc method, Psychiatric healthcare equity}

\begin{abstract}
A significant level of stigma and inequality exists in mental healthcare, especially in under-served populations. Inequalities are reflected in the data collected for scientific purposes. When not properly accounted for, machine learning (ML) models leart from data can reinforce these structural inequalities or biases. Here, we present a systematic study of bias in ML models designed to predict depression in four different case studies covering different countries and populations. We find that standard ML approaches show regularly biased behaviors. We also show that mitigation techniques, both standard and our own post-hoc method, can be effective in reducing the level of unfair bias. No single best ML model for depression prediction provides equality of outcomes. This emphasizes the importance of analyzing fairness during model selection and transparent reporting about the impact of debiasing interventions. Finally, we provide practical recommendations to develop bias-aware ML models for depression risk prediction.
\end{abstract}
\begin{document}
\flushbottom
\maketitle
\thispagestyle{empty}
\section*{Introduction}
Depression is a leading cause of disability worldwide, a major risk factor for the global burden of disease, and can even lead to suicide~\citep{Friedrich:2017ja,Bachmann:2018ij}. Taking into account that the global prevalence of depression increased by 25\% during the COVID-19 outbreak~\citep{Bueno-Notivol:2021ijchp}, being able to identify those individuals at risk would be of great value to enable the application of personalized preventive measures. To this end, it is necessary to characterize the factors leading up to the development of depression. Research to date points to the importance of both genetic and environmental factors (as well as their interaction) in the etiology of depression~\citep{Anttila:2018science,Geschwind:2015science}. Furthermore, environmental factors have been shown to co-occur, exerting cumulative effects on depression risk. The totality of these environmental influences is often referred to as the exposome and includes environmental and lifestyle factors, as well as traumatic life events~\citep{Harald:2017jaci}. Exposome data does not only provide an alternative picture, it is also relatively inexpensive and easy to acquire, typically through questionnaires~\citep{Olesen:2012ejn}. Motivated by the successful application of machine learning (ML) in different contexts of the medical domain, there is a spike in the use of ML for the detection, diagnosis, and treatment of depression~\citep{Chen:2019ama,Park:2021jamanetwork,Nemesure:2021sr}. Specifically, supervised ML methods are commonly used to learn predictive models from historical data, which are then applied to predict possible illness development in new cases and patients. 

Recently, concerns have been raised about algorithmic bias~\citep{Chen:2019ama} and the undesirable ability of ML models of amplifying unfair behaviors masked in past practice, that is, in the data used for model learning. The term “algorithmic bias” refers to differences in the predictive power of models when applied to different subgroups of the population. These differences are particularly worrying if they are found when the subgroups are determined according to some protected attribute such as ethnicity, sex, or age. The subgroup that is adversely impacted by the bias of the ML model is known as the unprivileged group, and the subgroup that is unfairly benefited is known as the privileged group. Although this undesirable behavior of ML models is nowadays well known, assessing the bias of ML models (and trying to mitigate it) is not a common practice in healthcare applications. Chen et al.~\citep{Chen:2019ama} examine an ML algorithm on psychiatric notes to predict 30-day psychiatric readmission regarding sex, ethnicity, and insurance type without addressing algorithmic bias. Park et al.~\citep{Park:2021jamanetwork} reduce bias for clinical prediction models of postpartum depression only associated with one protected attribute - binarized ethnicity (Black individuals and White individuals). In mental health, several unintentional discriminative behaviors have been detected, which could potentially be reproduced by ML models if they are reflected in the training data. Specifically, a lack of representation of the patient subgroups has been reported; for example, some ethnic groups do not use mental health services as much as others due to cultural stigma surrounding mental illness~\citep{Corrigan:2014pspi,Wong:2017rand}. Prior research shows that the prevalence or incidence of depression differs across sex subgroups; women are about twice as likely as men to develop depression during their lifetime~\citep{Albert:2015jpn}. In addition, manifest discrepancies in relevant factors such as lifestyle or nutritional habits between subgroups have also been reported~\citep{Lubin:2003jn}. These, and possibly other factors, can be rooted in the data which is used to learn ML models for depression prediction.

In this paper, we investigate how ML algorithms could perpetuate or reduce structural inequality and unfair bias learned from the data. We present a systematic analysis of algorithmic bias in ML models designed to predict the presence or absence of depression from environmental and lifestyle data, using four public datasets: LONGSCAN~\citep{Runyan:2014longscan}, FUUS~\citep{Tran:2017plosone}, NHANES~\citep{CDC:year}, and the UK Biobank (UKB)~\citep{Sudlow:2015plosmed}. We study unfair bias on protected attributes including demographic factors (sex, ethnicity, nationality), socioeconomic status (age, income, academic qualifications), and co-morbidities (cardiovascular disease (CVD), diabetes), and evaluate the interplay of model accuracy and fairness. For this study, protected attributes were agreed based on standard choices in the related literature. We analyze the ability of standard bias mitigation techniques to reduce the discrimination level of the models learned for our four case studies. The mitigation effect is measured as the performance difference, in terms of fairness and standard ML metrics, before and after performing bias mitigation. We have found unfair biases in the behavior of the models learned with standard ML techniques regarding several protected attributes in all the case studies. We also found, however, that mitigation techniques are effective in reducing discrimination levels. Our results suggest that bias monitoring is pertinent in the evaluation of ML-based predictive models in mental health and current mitigation techniques provide a powerful toolset to mitigate unfair algorithmic bias.
\section*{Methods}
In this section, we introduce and perform an initial descriptive analysis of the available data of our case studies (Sec.~\hyperref[sec:datasets]{Datasets}). In the following, we describe the predictive models tested and the five strategies we considered to mitigate bias (Sec.~\hyperref[sec:bias_mitigation_approaches]{Bias mitigation approaches}). We also explain both standard ML and fairness evaluation metrics used in our experiments (Sec.~\hyperref[sec:evaluation_metrics]{Evaluation metrics}).
\phantomsection
\subsection*{Datasets}
\label{sec:datasets}
This study uses four public datasets: LONGSCAN, FUUS, NHANES, and UK Biobank. We select these datasets to cover a spectrum of sizes, from small to large, and to showcase diverse methodologies for diagnosing depression. We aim to assess potential biases across different scales and diagnostic methods. These differences allow us to investigate how the size of the dataset (both in terms of the number of samples and input variables) influences bias in both plain ML models and those combined with debiasing techniques. All the participants and/or their carers provided written informed consent in LONGSCAN, NHANES, and UKB studies. This was not required in the case of FUUS according to the laws that regulate “non-interventional clinical research” in France~\citep{Tran:2017plosone}. While LONGSCAN and FUUS are datasets of late adolescents, NHANES and UKB have most subjects between 40 and 80 years of age. Supplementary Table 1 describes the protected attributes considered for each dataset, as they differ between datasets. The LONGSCAN, FUUS, NHANES, and UKB datasets have relatively equal proportions of male and female subjects. A higher prevalence of depression in women versus men is evident in the LONGSCAN, NHANES, and UKB datasets. No sex effect is found among college freshmen in the FUUS dataset, which is consistent with previous studies~\citep{Feng:2014plosone,Bayram:2008spe,Ovuga:2006wp}. Other protected attributes have a highly skewed distribution.

\textit{Participants and features}. Participants in LONGSCAN were enrolled from five United States' sites (South, East, Midwest, Northwest, and Southwest), with different selection criteria, representing varying levels of risk or exposure to maltreatment during the period spanning from 1991 to 2012. The LONGSCAN interview and questionnaire data were collected when target children were at ages 4, 6, 8, 12, 14, 16, and 18. Out of 1354 total participants, we kept the 67.3\% of children who completed an interview at age 18 years including depression outcomes, leading to 911 samples available for our study. Among these 911 individuals, there were 363 cases with depression at the age of 18, and 548 controls. The study design is depicted in Supplementary Figure 1: data from three different stages (early childhood, late childhood, teen) were collected to predict depression at age of 18. Up to 23 descriptive variables grouped as demographic variables, lifestyles variables, and adverse exposures variables, both time-invariant and repeatedly measured along these stages, were considered (see Supplementary Table 2). Data is available under request; its use for this study was approved by the National Data Archive on Child Abuse and Neglect (NDACAN). Participants in FUUS were undergraduate students who underwent a compulsory medical visit at the university medical service in Nice (France) between September 2012 and June 2013.  Among 4184 total participants, there are 528 cases with depression and 3656 controls. A total of 62 biomedical and demographic features were used including binary, ordinal and continuous variables (see Supplementary Table 3). Participants in NHANES provided data between 2005 and 2018 and were selected by random sampling of the American population. Among 36259 total participants, there were 3168 cases with depression and 33091 controls. A total of 86 features were used in our study, including demographic data, socioeconomic status, medical history, lifestyle characteristics, and prescription medications. Factors that might involve data leakage, such as depression-specific medications, were excluded (see Supplementary Table 4). FUUS and NHANES datasets are publicly available. Participants in UKB were enrolled from 22 United Kingdom's centers from 2006 to 2010. Among the 461,033 participants initially without depression, 18112 cases (3.93\%) developed depression. Up to 143 descriptive variables were considered, including demographic data, socioeconomic status, medical history, lifestyle characteristics, early life factors, and traumatic events (see Supplementary Table 5). Data is available under request; its use for this study was approved by UKB, under the project title “Association between Early-Life-Stress and Psycho-Cardio-Metabolic Multi-Morbidity: The EarlyCause H2020 Project” (application number 65769). 

\textit{Building the ground truth - depression outcome}.  We acquired ground truth label information –whether a participant has depression or not– using dataset-specific information. In LONGSCAN, depression was assessed using a self-reported questionnaire at age 18, which includes a specific question regarding having depression. In NHANES, the Patient Health Questionnaire-9 (PHQ-9) was used. This screening questionnaire consists of 9 items (scored 0-3) and has a specificity and sensitivity of 88\% for major depressive disorder (MDD) at a threshold score of 10 or more~\citep{Kroenke:2001jgim}. Therefore, we chose the threshold at PHQ-9 score 10. Additionally, we carefully excluded participants' feelings and expressions, as well as lifestyle characteristics (physical activity, diet, sleep habits) from the set of descriptive features, which could cause label leakage. In FUUS, the depression outcome was evaluated in a two-stage process. If the result of an initial 4-item screening questionnaire indicated possible presence of MDD (at least two of the four symptoms present), the participants were assessed by a medical provider for the full Diagnostic and Statistical Manual of Mental Disorders Fourth Edition (DSM IV) criteria~\citep{Nemesure:2021sr}. In UKB, the depression outcome was defined as an occurrence of a depressive episode (ICD10 code F32 and F33) after the date of assessment, which was drawn from hospital inpatient diagnoses or conditions self-reported. Note that this label information might be noisy due to the use of questionnaires. Addressing biased labels is a complex ML problem that falls beyond the scope of our study.  

Two popular types of ML models were learned from this data: (i) logistic regression (LR)~\citep{Yu:2011ml}, a linear classifier,  and (ii) extreme gradient boosting (XGB)~\citep{Chen:2016xgboost}, a boosting ensemble method. In this paper, we do not focus on the algorithmic aspects of the ML methods considered, but rather on their clinical application and the fairness assessment of their predictions. We use \(k\)-fold cross validation (\(k=5\) for UKB, \(k=10\) for the rest) for performance evaluation and nested cross-validation for hyper-parameters tuning (see Supplementary Table 6). Tukey's range test~\citep{Tukey:1949biometrics} identifies the 95\%-significance for true-positive rates for each group. 
 
\phantomsection
\subsection*{Bias mitigation approaches}
\label{sec:bias_mitigation_approaches}
Many techniques have been proposed over the last few years to address algorithmic fairness~\citep{Xu:2022ebiom}. However, there is a significant shortfall in addressing fairness and bias concerns when ML has been applied to the field of psychiatry. Only a handful of studies have adopted methods to counteract bias. For instance, reweighing (RW) bias-mitigation technique~\citep{Calders:2009icdmw} was used to minimize bias when forecasting future benzodiazepine administrations~\citep{Mosteiro:2022info}. Likewise, others applied Suppression (SUP)~\citep{Verma:2018fairness} and RW approaches to reduce bias in the prediction of postpartum depression~\citep{Park:2021jamanetwork}. In healthcare and other fields, the Disparate Impact Remover (DIR) method~\citep{Feldman:2015kdd} has shown its ability to effectively mitigate bias while maintaining a satisfactory level of predictive performance. Furthermore, the Calibrated Equalized Odds Post-processing (CPP) method~\citep{Pleiss:2017nips} was proposed in the context of healthcare to predict whether an individual has a heart condition. Among all the available bias-mitigation techniques, we consider four standard methods in this study: SUP, RW, DIR, and CPP. Moreover, we propose a novel post-hoc disparity mitigation named Population Sensitivity-Guided Threshold Adjustment (PSTA). 

\textit{Suppression (SUP)}. Protected attributes are directly removed from the training dataset under the assumption that access to this information is the main cause of bias. First, the protected attribute is removed from the dataset. Then, a new new ML model is learned from this new version of the dataset. 

\textit{Reweighing (RW)}. This method weights the samples in each (group, label) combination differently to make the protected attribute and outcome statistically independent of each other before model learning. The weight of a sample is directly proportional to the frequency of its label in the whole population and inversely proportional to the frequency of its label in its subgroup. The model is learned from this new training dataset with weighted samples.

\textit{Disparate impact remover (DIR)}. Given a dataset $D = (X, Y, C)$, with protected attribute $X$, remaining attributes $Y$, and class variable or outcome $C$, this repair process attempts to remove bias in the remaining features $Y$. New values are assigned to all the cases and variables in $Y$. The new values ensure that all the groups follow the same distribution over every variable, making adjustments based on percentiles and quantile functions. The predictive model is learned from the new training dataset. In the three previous bias-mitigation techniques, we act on the training data. Once a newly prepared version of the training dataset is obtained, an ML model is learned from it, with the effectiveness of bias mitigation evaluated based on this model's outputs. The following two techniques perform differently. The model is learned from the original data. Then, the outputs of the model are modified under some criteria to reduce disparities. The success of the mitigation approach is then evaluated using these modified outputs.

\textit{Calibrated equalized odds postprocessing (CPP)}. The method calibrates the predicted probability so that the false-positive rate or the false-negative rate of privileged and unprivileged groups are on average equal. It modifies the score outputs of the model for the different subgroups so that the output labels meet an equalized odds objective. In our clinical application, we focus on identifying individuals at risk of depression and desire equitable outcomes; therefore, we consider recall  more important than precision, which leads us to set a cost constraint objective: the equal false-negative rates between the subgroups.

\textit{Population Sensitivity-Guided Threshold Adjustment (PSTA)}. This approach, our own bias mitigation technique, is inspired by the observation that when there are prevalence rate discrepancies between groups in the training dataset, ML models usually associate the positive class mainly with the characteristics of the subgroup of highest prevalence. Supplementary Figures 2a and 2b
illustrate the predicted probability distributions generated by a ML model with data from UKB with different prevalences for females and males suffering from depression. The distributions differ significantly. The difference arises from the prevalence rate discrepancies in the training dataset, leading the model to inherently assign lower probabilities of suffering from depression to individuals from the male subgroup, which has a lower prevalence rate. Applying the default single decision threshold at 0.5 for both groups results in a lower TPR and a higher FPR for the male group, exacerbating disparities in the outcomes. This example model would benefit from a tailored threshold adaptation that considers the disparities in the distribution of probabilities between different subgroups, ultimately improving both fairness and accuracy. 
The use of threshold adaptation to address bias and unfairness in ML models is not entirely new, as evidenced by previous works such as~\citep{Rodolfa2020,Jang2022}. However, our proposal introduces a novel procedure for obtaining the optimal threshold, which differentiates it from existing approaches. Our method aims to improve fairness while maintaining model performance. It is guided by the fairness metric Equal Opportunity, which is applicable to categorical protected features, i.e., it is not limited to binary features only. The main objective is to enhance sensitivity for unprivileged groups while maintaining an acceptable FPR, which refers to a level that is not significantly higher than the overall population's FPR, ensuring a balance between minimizing incorrect positive predictions and effectively detecting true positive cases. It determines subgroup-specific thresholds for unprivileged groups to ensure that their sensitivity aligns with that of the overall population on training data, encompassing the sensitivity of all subgroups. For the remaining groups, the standard threshold at 0.5 is employed. This conventional 0.5 threshold can be adjusted as a hyperparameter to better suit specific context requirements. The whole procedure is thoroughly detailed in Algorithm~\ref{alg:PSTA}. As depicted in Supplementary Figure 2c
, by utilizing the optimal threshold generated by PSTA on the training set for the unprivileged group (male in this case), the FPR of the unprivileged group reaches an acceptable value in the test set. This example is illustrative of the type of correction that PSTA enforces, leading to more accurate and fair predictions. 

In this study, we use three different pre-processing mitigation techniques (SUP, RW, and DIR), which operate over training data, and two different post-processing mitigation techniques (CPP and PSTA), which operate over the model’s predictions. See a graphical description of how they operate in Supplementary Figure 3. Generally, pre-processing and post-processing techniques are model-agnostic. There exists a third type of bias mitigation technique, called in-processing methods, which operate during classifier construction, leading to a significant reliance on the specific type of model in use. Given our study's focus on broadly applicable and adaptable mitigation techniques, we concentrated on pre- and post-processing methods, as their adaptability aligns better with our research objective. An in-depth comparison of bias mitigation techniques is beyond the scope of this study, which aims to demonstrate that with a diverse toolkit, ML practitioners can expect to build a model with increased fairness without a significant loss of predictive performance.
\algrenewcommand\algorithmicrequire{\textbf{Input:}}
\algrenewcommand\algorithmicensure{\textbf{Result:}}
\begin{algorithm}
\caption{Population Sensitivity-Guided Threshold Adjustment (PSTA)}
\label{alg:PSTA}
\begin{algorithmic}[1]
\Require A training set \(D=\{({y_i, x_i, c_i})\}_{i=1}^n\), where \( y_i \) are the values of the descriptive variables, \( x_i \) is the value of the protected attribute and \( c_i \) the class for individual \( i \), and \( P = \{ p_i \}_{i=1}^n \) where \( p_i \) is the vector of probabilities predicted for individual \( i \) and each class label.
\Ensure A vector of threshold values \(\mathbf{\theta}\), one per group \(\mathbf{\theta} = (\theta_x)_{x \in \Omega_X}\)
\State Initialize threshold vector to \(\boldsymbol{\theta_x} = 0.5\) for all the groups, \(x \in \Omega_X\)
\State \(sensitivity_{\text{overall}}\) := Calculate sensitivity in the overall training set with current \(\mathbf{\theta}\), i.e., all decision thresholds at \(0.5\)
\For {each \emph{unprivileged} group, \(u\) in \(\Omega_X\)}
    \State Initialize minimum difference \(minDiff = \infty\)
    \State Initialize optimal threshold for group \(\hat{\theta}_u = 0.5\)
    \For {each possible threshold value \(\theta' \in [0,1]\)}
        \State Calculate the sensitivity of the unprivileged group \(u\), \(sensitivity_u\)
        \State \(diff := \text{abs}(sensitivity_u - sensitivity_{overall})\)
        \If {\(diff < minDiff\)}
            \State \(minDiff = diff\)
            \State \(\hat{\theta}_u = \theta'\)
        \EndIf
    \EndFor
    \State Adopt threshold \(\theta_u = \hat{\theta}_u\)
\EndFor
\State Use the vector of thresholds \(\mathbf{\theta}\) to predict the class of new data instances
\end{algorithmic}
\end{algorithm}
\phantomsection
\subsection*{Evaluation metrics}
\label{sec:evaluation_metrics}
\textit{Performance metrics}. To assess predictive performance, we consider two standard ML performance metrics, namely, the area under the receiver operating characteristic curve (AUC-ROC), that is an average of false positive rates versus true positive rates across all the possible thresholds, and the balanced accuracy (BAcc), which is the arithmetic mean of sensitivity and specificity, using the presence and absence of depression as the positive and negative class, respectively. This weighting is especially beneficial for extremely imbalanced datasets, ensuring equitable representation in performance assessment. When discussing the fairness-accuracy trade-off, we report on empirical accuracy measured by the latter metric because, in practice, classification will be performed at a fixed threshold~\citep{Park:2021jamanetwork}, making BAcc a more pertinent measure in such scenarios. We also report AUC-ROC performance after applying the most effective bias mitigation algorithms in Supplementary Table 7.

\noindent \textit{Fairness metrics}. There are three kinds of fairness metrics depending on the fairness concept that they account for: metrics that account for individual fairness, for group fairness, and for both~\citep{Bellamy2018}. In this study, we focus exclusively on group fairness metrics for binary classification tasks, which means that subgroups defined by a protected attribute should receive similar model outcomes measured by some statistical metrics. In this clinical application, identifying all individuals at risk of depression is desirable; therefore, we consider sensitivity over specificity which leads us to compare true-positive rates across subgroups. Specifically, we choose the equal opportunity difference (EOD) metric, which states that a binary classifier is fair if its true-positive rates (TPR) are equal across groups (i.e., a value of 0 indicates complete fairness). Let \( D = (X, Y, C) \) be the dataset, with the protected variable \( X \), regular descriptive variables \( Y \), and the class variable \( C \). Predictions provided by an ML model are denoted \( \hat{C} \). Let us define \( \Omega_X \) as the set of possible values of variable \( X \). For example, for the protected attribute “sex”, \( \Omega_X = \{\text{male, female}\} \). A subgroup of the population is formally defined as all the samples in dataset \( D \) with the same value \( x \in \Omega_X \) assigned to the protected attribute \( X \). We define the TPR of a specific subgroup \( x \in \Omega_X \) as
\begin{center}
\( TPR_x(\hat{C}) := \mathbb{E}_{Y}[\hat{C} | C = 1, X = x] \)
\end{center}
Exact equality, \( TPR_{x_0}(\hat{C}) = TPR_{x_1}(\hat{C}) \), for \( x_0, x_1 \in \Omega_X \), is often hard to verify or enforce in practice. More generally, we use differences among the TPRs of the different subgroups \( x \in \Omega_X \) to measure the level of discrimination \( \Gamma \)~\citep{Chen:2018nips}:
\begin{center}
\( \Gamma^{\text{TPR}}(\hat{C}) := | \text{TPR}_{x_0}(\hat{C}) - \text{TPR}_{x_1}(\hat{C}) | \)
\end{center}
Specifically, EOD measures the difference of TPR between the unprivileged and privileged groups:
\begin{center}
\( \text{EOD} = \text{TPR}_{x=\text{unprivileged}} - \text{TPR}_{x=\text{privileged}} \)
\end{center}
\noindent This study is contextualized in a project focused on developing efficient screening and risk prediction tools for depression in primary care settings. At this point, underdiagnosis has more severe implications than misdiagnosis. When a misdiagnosis occurs, patients still receive clinical care, and clinicians can draw upon additional symptoms and data sources to correct the error. On the other hand, underdiagnosis may lead to individuals not receiving the necessary treatment and support, exacerbating their mental health condition. This reasoning motivates our decision for the group fairness EOD metric, in line with previous applications in the mental health domain~\citep{Park:2021jamanetwork,Mosteiro:2022info}. EOD is an attainable and practical fairness criterion which mandates equal TPRs across the demographic subgroups. This decision allows us to focus on minimizing disparities in TPRs across different groups, ensuring that all individuals at risk of depression are identified and provided with the necessary care and support they require.
\section*{Results}
In this section, we study the behavior of the diﬀerent ML models before and after bias mitigation techniques are applied. Firstly, ML models are learned from data and their predictive performance and unfair bias are quantified (Sec.~\hyperref[sec:initial_model_performance_and_bias_assessment]{Initial model performance and bias assessment}). Secondly, bias mitigation techniques are applied to these models and to assess their efficiency, we evaluate the interplay between predictive performance and fairness in the adapted models (Sec.~\hyperref[sec:model_performance_after_bias_mitigation]{Model performance after bias mitigation}). We perform this analysis in four datasets for the prediction of depression: LONGSCAN, FUUS, NHANES, and UKB. 
\phantomsection
\subsection*{Initial model performance and bias assessment}
\label{sec:initial_model_performance_and_bias_assessment}
\begin{figure}[!htb]
    \centering
    \includegraphics[width=\textwidth, height=0.75\textheight]{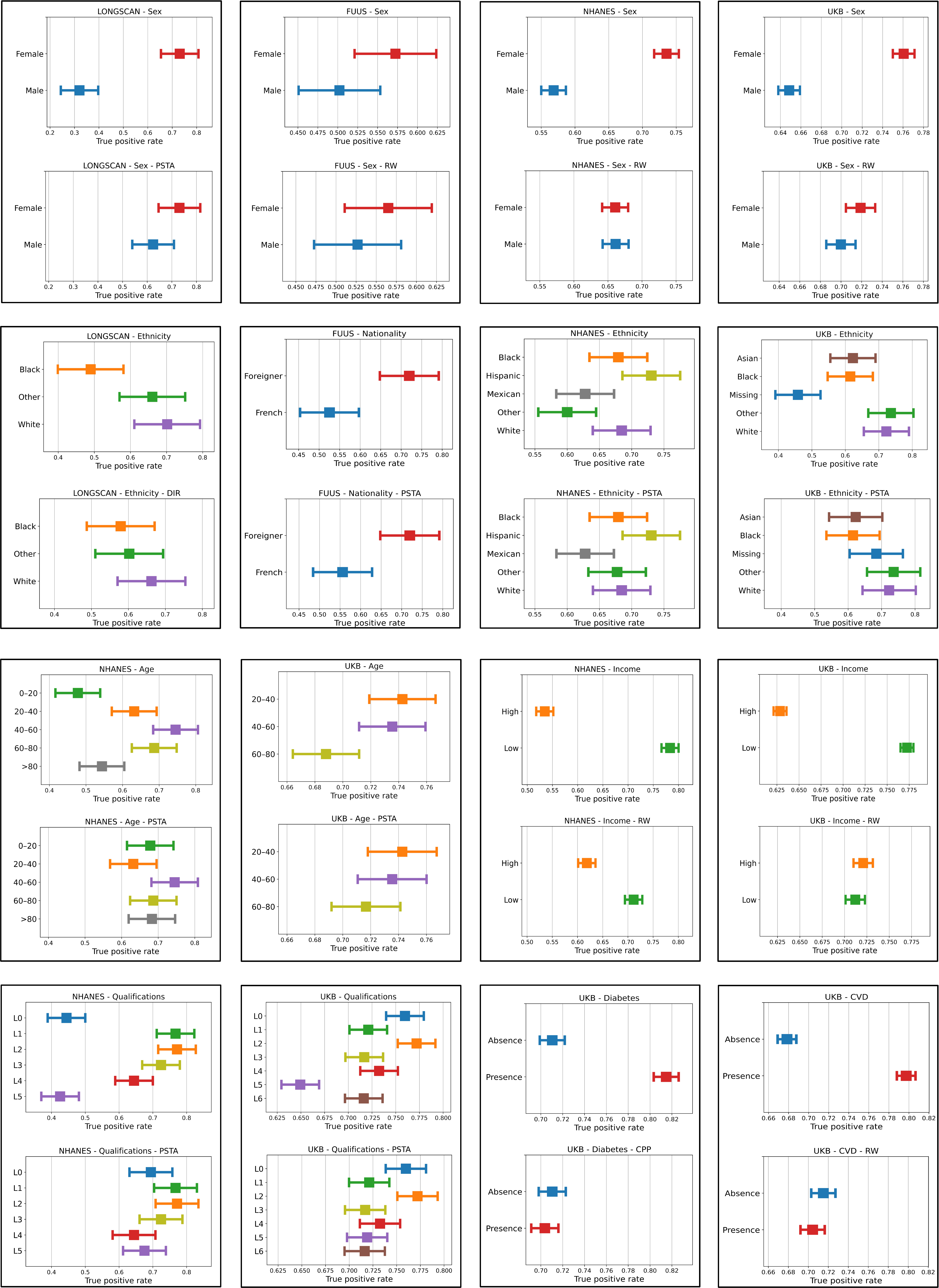}
    \captionsetup{font=small} 
    \caption{Comparative analysis of group-specific TPRs for LR classifiers: baseline and bias mitigation outcomes. Each plot represents a Dataset-Protected attribute pair, with paired rows displaying base classifiers and debiased classifiers, reporting the best results among the tested bias mitigation algorithms. Note that the base classifiers show regularly biased behaviors due to a lack of representation, varying rates of depression across groups, unequal distribution of features between groups, or a combination of any of these characteristics in the four different study populations.}
    \label{fig:comparative_analysis}
\end{figure}

For the sake of simplicity, we only report results with LR models in this manuscript. Results with XGB models, qualitatively similar to those of LR, are available in the Supplement. Regarding predictive performance, LR models achieve lower prediction accuracy with LONGSCAN  and FUUS datasets, with BAcc of 0.621 (95\% CI: 0.577-0.664) and 0.615 (95\% CI: 0.598-0.632), respectively. The predictive performance of LR models is considerably better with NHANES and UKB datasets, with BAcc of 0.719 (95\% CI: 0.711-0.727) and 0.729 (95\% CI: 0.725-0.734), respectively. For a comprehensive performance report including AUC-ROC metrics, see Supplementary Table 8. These results support that the predictive ability of ML models increases with the amount of training data, as NHANES and UKB are larger. Figure \ref{fig:comparative_analysis} displays the fairness metric of the base and debiased LR models by dataset, protected attribute and subgroup, providing a comprehensive comparison of fairness performance before and after the application of bias mitigation techniques. In each plot, the horizontal shift or difference in performance for different subgroups reveals the bias. In each frame, the upper plot shows results before applying a bias mitigation technique, and the lower plot shows results after mitigation. Let us analyze the bias assessment by (type of) protected attribute: demographic factors, socioeconomic status, and co-morbidities.

\textit{Protected attributes: demographic factors (sex, ethnicity, nationality)}. We show consistent sex differences across datasets in the top row of Figure \ref{fig:comparative_analysis}, with higher TPRs for female subjects. The differences in TPRs between sexes are only statistically significant at the 95\% confidence level for the LONGSCAN, NHANES, and UKB datasets. Note that there are no sex differences in rates of depression among subjects in the FUUS dataset (see Supplementary Table 1). Interestingly, the mean difference between sexes in the UKB and NHANES datasets (0.1121, 0.167, \(p < 0.001\)) is less than in the LONGSCAN dataset (0.4103, \(p < 0.001\)). Sex, the top importance feature in LR and XGB for the LONGSCAN dataset, has a much lower rank in the feature importance ranking for the NHANES and UKB datasets (see the feature importance rankings in the Supplement), which means depression outcome is less sensitive to sex bias in the NHANES and UKB datasets with a large sample size and features compared to the LONGSCAN dataset. We highlight the benefit obtained from considering a larger number of risk factors in the predictive model and adequate sample size to reduce bias. This evidence is in line with~\citep{Chen:2018nips}, where enhanced data collection is pointed out as a means to lessen discrimination without sacrificing accuracy. The second row of Figure \ref{fig:comparative_analysis} shows differences in TPRs between racial groups, which were not statistically significant, with black subjects and ``other/multiracial'' subjects having the lowest true-positive rates for LONGSCAN and NHANES datasets, respectively; except for the case between black subjects and white subjects in the LONGSCAN dataset, their TPRs have non-overlapping confidence intervals, indicating a significant difference. Interestingly, the UKB subjects in the ``do not know/prefer not to answer'' (‘Missing’) group have the lowest TPR, compared with others. As shown in Figure \ref{fig:comparative_analysis}, TPRs for nationality have non-overlapping confidence intervals. Specifically, foreigner subjects have a higher rate than French subjects. We note that foreigner subjects have a higher observed depression rate in the training set. 

\textit{Protected attributes: socioeconomic status (age, income, academic qualifications).} We find that the TPRs do not differ much across age groups with many overlapping intervals. However, subjects under 20 years old have the lowest TPR. This may be partially due to the fact that small subset sizes (see Supplementary Table 1) may not reflect accurate depression rates amongst the adolescents and young adults subpopulation, whose symptoms of depression and other mental illnesses have increased significantly over the last decades~\citep{Twenge:2019jap}. Differences in TPRs in the NHANES dataset are also observed between qualification groups. As seen in Figure \ref{fig:comparative_analysis}, subjects in the ``Level 5'' group have the lowest TPR, compared with others in both the NHANES and UKB datasets. Subjects in the ``Level 0'' (Refused/Don’t know/Missing) group in the NHANES dataset are also on the unprivileged side. As shown in Figure \ref{fig:comparative_analysis}, TPRs for income have non-overlapping confidence intervals. Specifically, low-income subjects have a much higher rate than high-income subjects. We note that low-income subjects have a higher baseline observed event rate.

\textit{Protected attributes: co-morbidities (CVD, diabetes)}. As shown in Figure \ref{fig:comparative_analysis}, TPRs for CVD and diabetes all have non-overlapping confidence intervals. Specifically, individuals experiencing CVD/diabetes have a much higher rate than subjects without CVD/diabetes. These inequitable outcomes support that CVD and diabetes should be considered as important comorbidity of depression. 

\begin{figure}[!htb]
    \centering
    \includegraphics[width=\textwidth, height=0.5\textheight]{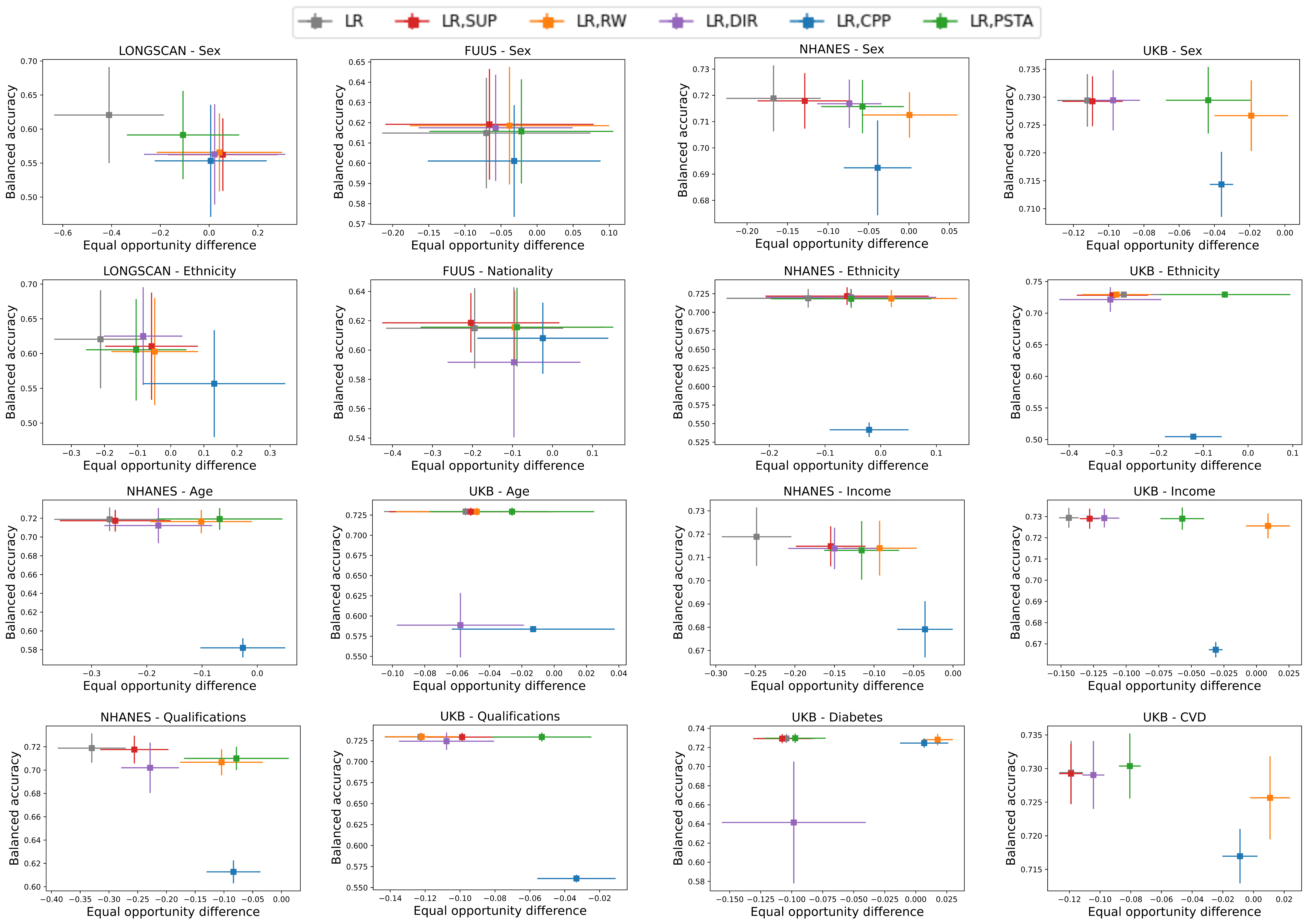}
    \captionsetup{font=small} 
    \caption{Fairness-accuracy performance in terms of EOD vs. BAcc of the base model and the new classifiers after applying five bias mitigation algorithms to LR classifiers. Each plot shows the results per subgroup for a Dataset-Protected attribute pair.}
    \label{fig:Figure2}
\end{figure}
\phantomsection
\subsection*{Model performance after bias mitigation}
\label{sec:model_performance_after_bias_mitigation}
To assess bias mitigation, we analyze changes in fairness metrics between the previously discussed base models and the new classifiers obtained after bias mitigation. Moreover, given the clinical application, most people would not find it fair to reduce discriminatory outcomes if it identifies fewer actual positives overall. There exists an open discussion in the related literature~\citep{Calders:2009icdm,Menon:2018fat} regarding the actual existence of the so-called fairness-accuracy trade-off when bias mitigation is implemented, meaning that predictive performance is reduced if one tries to make the model fairer. A trade-off of accuracy for fairness is often undesirable in healthcare. Thus, we have two objectives: increasing accuracy and non-discrimination.  We report the results using 2D points combining two metrics: (i) a fairness metric, measured by EOD and (ii) a standard ML performance metric, measured by BAcc, on the test set, to determine whether our models for depression prediction are experiencing a fairness-accuracy trade-off. A model can be considered as fair if EOD is between -0.1 and 0.1, its ideal value is 0 and for BAcc, the larger the better. Figure \ref{fig:Figure2} presents, for each dataset and protected attribute, the 2D points in the fairness-accuracy space achieved by each bias mitigation technique. The base model, without bias mitigation, is also shown (gray point and cross). The implemented bias methods help to improve fairness, for all protected attribute perspectives. This is a desirable mitigation result. Note that Figure \ref{fig:comparative_analysis} showcases group-specific TPRs for LR classifiers after applying the best-performing bias-mitigation techniques for each protected attribute. This allows for a detailed evaluation of their effectiveness across different study populations. It is noteworthy that in general debiasing through RW, DIR, and PSTA substantially improves fairness without compromising model accuracy. SUP only partially achieves fairness between groups. This method removes the protected attribute from the training dataset, as they are considered biased features. This result suggests that bias is not only contained in those features but elsewhere. DIR not only excludes these attributes but also adjusts non-protected features that are highly correlated to them. In any case, this repair tool is a good baseline to investigate whether this bias comes from non-protected features. On the other hand, the CPP technique preserves precision when calibrating recall, which results in BAcc reduction, according to most of the protected attributes, except for the diabetes attribute. Furthermore, it is important to highlight that, when applying bias-mitigation methods to the LONGSCAN dataset with the protected attribute sex, a significant decrease in BAcc is observed across all methods. This observation serves as evidence that the model may struggle to effectively learn the underlying structure of the data due to the relatively small dataset. Consequently, this emphasizes the necessity of having sufficient samples and relevant predictors when utilizing ML algorithms for risk prediction tasks in order to develop both fair and accurate models. Note that TPRs for the best mitigation technique (RW) regarding the “income” protected attribute on models learned from the NHANES dataset still have non-overlapping confidence intervals, although the mean difference between groups was considerably reduced from 0.2485 (p $< 0.001$) to 0.0929 (p $< 0.001$).
\section*{Discussion}
ML algorithms have achieved state-of-the-art performance in many clinical tasks. However, to deploy them in these life-or-death-stakes applications, it must be understood that they can induce biases against unprivileged subgroups and precautionary actions need to be taken in different deployment stages~\citep{deHond:2022npjdm}. Here, we leverage our empirical study on four datasets to analyze the need of bias mitigation techniques, as well as their effectiveness, when using ML models to predict mental health issues such as depression.

\textit{Bias found when following the standard ML approach}. Our results indicate that models learned following the standard ML approach show regularly unfair biased behaviors (see Figure~\ref{fig:comparative_analysis}). We find that, for the classification problem, unequal distribution of classes between groups in the training dataset can lead the predictive model to learn that one group has a higher probability of being part of one class or another. (In Supplementary Table 1, evidence of this behavior can be observed). Therefore, ML models trained on the unbalanced dataset of a trial population, even if the sample in clinical trials is representative of the patient population, provide potential inequitable outcomes if deployed without fairness analysis. This evidence encourages ML practitioners in healthcare not only to report the model performance on the overall population regardless of the subject membership to subpopulations but also to audit and address algorithmic bias. 

\textit{Bias can be mitigated}. Our results show that the bias mitigation techniques improve fairness compared to the no-intervention base models. All the techniques considered enhance results in terms of the difference in TPRs, in diﬀerent proportions. However, the techniques exhibit diﬀerences regarding the eﬀect on the accuracy of the classifiers. Those learned in combination with the RW, DIR, and PSTA mitigation techniques tend to preserve predictive performance, whereas other techniques (SUP, CPP) usually compromise the level of accuracy (see Figure~\ref {fig:Figure2}). Our findings point out that the RW technique combined with the use of a larger number of risk factors diminishes the impact of the protected attributes (and other possible proxy attributes) on the outcome of the model, which leads to reduced algorithmic bias in NHANES and UKB datasets. This solution is appropriate in our case study as it performs well when it is integrated into predictive model learning while preserving the distribution and values of the original training data, unlike DIR and SUP. In addition, we find that our proposed post-hoc disparity mitigation method (PSTA) tends to mitigate bias while preserving predictive performance. In this method, the distinct treatment of subpopulations aims to address subgroup-specific disparities. Conventional one-size-fits-all methods might inadvertently accentuate biases, especially for underrepresented groups. Therefore, our approach ensures fairness by recognizing and addressing these unique subgroup challenges. However, as other post-doc technical solutions for imposing fairness, it might be difficult to obtain the optimal threshold for subgroups with small populations underrepresented in the training set. The choice of the method must depend on the specific domain of application and desired outcomes. In the larger effort to create a fair system overall, methods like RW and DIR not only increase the TPR for the unprivileged groups but also reduce the privileged group's FPR, resulting in fewer false positives. In contrast, PSTA increases the TPR of the unprivileged groups, accepting a higher FPR to reduce the TPR gap without intervening in the privileged groups. In primary care settings, where early detection and prevention are key, PSTA may be more appropriate, as it ensures that more people are identified for further evaluation and potential intervention, maximizing the overall benefit. Conversely, in secondary care settings focused on diagnosing and treating specific conditions, techniques such as RW and DIR may be more suitable, as they balance TPR and FPR across demographic groups, ensuring effective and fair resource allocation. We also highlight the importance of prioritizing adequate data collection before employing any debiasing technique to improve fairness in predictive models. By doing so, we aim to encourage communities to open health data, further contributing to the development of more equitable ML solutions. 

\textit{Fairness-accuracy, a real trade-off?} In fairness literature, the existence of a trade-off between fairness and accuracy is a common assumption, that is, that fairness cannot be improved without sacrificing predictive performance~\citep{Calders:2009icdmw}. A few studies also have pointed out that this trade-off may necessitate the application of more complex methods~\citep{Friedler:2019fat,Zafar:2017aistats}. Based on our empirical observations, we find that the fairness–accuracy trade-off for the datasets examined in this paper can be bent if a set of bias mitigation techniques is considered, and not just one. Standard techniques RW, DIR, and CPP, or our proposed method PSTA can reduce bias while preserving predictive performance for specific (dataset, protected attribute) pairs. This was particularly evident for the modest-sized to large datasets (FUUS, NHANES, UKB), though the dynamics shifted a bit for the smaller dataset (LONGSCAN). Thus, this trade-off is not consistently observed in our case studies in depression prediction without requiring complex ML methods. In line with~\citep{Rodolfa:2021nmi}, this evidence encourages the ML community to intentionally propose frameworks that maximize both predictive performance enhancement and bias reduction, aiming to bend the trade-off. There is probably no golden bullet, as there is no single best ML model for all prediction problems providing equality of outcomes naturally. ML practitioners need to figure out which combination of type of classifier and bias mitigation algorithm is appropriate for the use case at hand so that it produces the best results in terms of both accuracy and fairness. 

In conclusion, we conduct an empirical study on four exposome datasets to show the ability of bias mitigation techniques to increase fairness of machine learning models obtained to predict depression from environmental and lifestyle data. In addition, our main eﬀort in this work has been directed toward providing empirical evidence to encourage clinical decision-makers to carefully evaluate a proposed framework in terms of both its accuracy and fairness prior to deployment. Experimental results support the idea that it is possible to improve algorithmic fairness without sacrificing their predictive performance. We consider that our promising results could enable a wider use of ML techniques in mental healthcare. This should inevitably go hand-in-hand with the assessment of possible biases in the models and the appropriate mitigation techniques if required. In the future, we expect to examine simultaneously the eﬀects of having multiple protected attributes such as ethnicity, sex, socio-economic status, geographical location, or co-morbidities (type 1 and type 2 diabetes, or cardiovascular disease),  as well as extending to use protective factors data, such as intelligence, temperament, cognitive appraisal, support from a significant person, which may counteract the negative eﬀects of risk factors for depression, along with environmental and lifestyle data. In addition, we aim to integrate genetic and biological data, which are robust risk factors for depression, in our research. 

\section*{Data Availability}
Code for data processing and analysis is available at \url{https://github.com/ngoc-vien-dang/FairML-Depression}. Detailed dataset sources and accessibility conditions are also provided in the repository's README.

\bibliographystyle{plainnat}

\section*{Acknowledgements}
VND, RHM, CC, JHG, and KL have received funding from the European Union’s Horizon 2020 research and innovation programme under Grant Agreement No 848158, EarlyCause.
\section*{Author contributions statement}
Vien Ngoc Dang: Methodology, Investigation, Data curation, Software, Visualization, Writing - original draft, review \& editing. Anna Cascarano, Rosa H. Mulder, Charlotte Cecil, and Maria A. Zuluaga: Writing - review \& editing. Jerónimo Hernández-González: Conceptualization, Writing - original draft, review \& editing, Co-supervision. Karim Lekadir: Conceptualization, Resources, Writing - review, Supervision.

\section*{Additional information}
\textbf{Competing interests}
The authors declare that they have no known competing financial interests or personal relationships that could have appeared to influence the work reported in this paper.
\end{document}